\documentclass{article}
\usepackage{spconf,amsmath,graphicx}
\usepackage{makecell}

\usepackage{enumitem}
\setlist{nosep, leftmargin=14pt}
\usepackage{multirow}

\usepackage{mwe} 

\title{DINOv3-Guided Cross Fusion Framework for Semantic-aware CT generation from MRI and CBCT}

\name{
\vspace{-1pt}
  \begin{tabular}{c}
    Xianhao Zhou$^1$, Jianghao Wu$^2$, Ku Zhao$^1$, Jinlong He$^1$, \\
    Huangxuan Zhao$^3$, Lei Chen$^4$, Shaoting Zhang$^{1,5}$, Guotai Wang$^{1,5}$\thanks{Corresponding author: Guotai Wang (guotai.wang@uestc.edu.cn).}
  \end{tabular}
}
\address{$^1$ School of Mechanical and Electrical Engineering,\\ 
University of Electronic Science and Technology of China, Chengdu, China \\
$^2$ Faculty of Information Technology, Monash University, Melbourne, Australia \\
$^3$ Institute of Artificial Intelligence, School of Computer Science, Wuhan University, China \\
$^4$ Department of Radiology, Union Hospital, Tongji Medical College, \\ 
Huazhong University of Science and Technology, Wuhan, China \\
$^5$ Shanghai Artificial Intelligence Laboratory, Shanghai, China \\
}

\begin{document}
%

\maketitle

\begin{abstract}
Generating synthetic CT images from CBCT or MRI has a potential for efficient radiation dose planning and adaptive radiotherapy. However, existing CNN-based models lack global semantic understanding, while Transformers often overfit small medical datasets due to high model capacity and weak inductive bias. To address these limitations, we propose a DINOv3-Guided Cross Fusion (DGCF) framework  that integrates a frozen self-supervised DINOv3 Transformer with a trainable CNN encoder-decoder. It hierarchically fuses global representation of Transformer and local features of CNN via a learnable cross fusion module, achieving balanced local appearance and contextual representation. Furthermore, we introduce a Multi-Level DINOv3 Perceptual (MLDP) loss that encourages semantic similarity between synthetic CT and the ground truth in DINOv3's feature space. Experiments on the SynthRAD2023 pelvic dataset demonstrate that DGCF achieved state-of-the-art performance in terms of MS-SSIM, PSNR and segmentation-based metrics on both MRI$\rightarrow$CT and CBCT$\rightarrow$CT translation tasks. To the best of our knowledge, this is the first work to employ DINOv3 representations for medical image translation, highlighting the potential of self-supervised Transformer guidance for semantic-aware CT synthesis. The code is available at https://github.com/HiLab-git/DGCF.

\end{abstract}


\section{Introduction}
In modern radiotherapy planning and diagnostic workflows, Computed Tomography (CT) remains the standard imaging modality due to its reliable Hounsfield Unit (HU) mapping and quantitative tissue characterization~\cite{sct_review}. However, acquiring an additional planning CT introduces extra radiation exposure and logistical burden~\cite{sct_review}. Meanwhile, Cone-Beam CT (CBCT) and Magnetic Resonance Imaging (MRI) have become increasingly accessible for treatment guidance and monitoring, offering an opportunity to generate synthetic CT (sCT) images directly from these modalities, especially in the challenging pelvic region, to reduce dose and improve workflow efficiency~\cite{sct_review, cbct2sct_review}.

Recent advances in deep learning have enabled remarkable progress in medical image-to-image translation, particularly in CBCT$\rightarrow$CT and MRI$\rightarrow$CT synthesis~\cite{sct_review, cbct2sct_review}. However, effectively combining global semantic understanding with local structural fidelity remains difficult. Convolutional Neural Networks (CNNs), while strong at modeling fine-grained textures, struggle to capture long-range dependencies and contextual semantics~\cite{cnn_and_Transformer_comparison}. In contrast, Transformers have strong global modeling capabilities but, due to their large model size, high data requirements, and lack of inductive biases, they are prone to overfitting on small medical datasets~\cite{cnn_and_Transformer_comparison, inductive_bias}. To overcome this data-efficiency bottleneck, the community increasingly turns to foundation models pretrained on billion-scale natural-image collections ~\cite{fundation_models}. Among the latest advances, DINOv3~\cite{dinov3} has shown powerful and generalizable feature extraction across a variety of downstream tasks, including segmentation and detection~\cite{segdino,deimv2}. Nevertheless, its potential in generative tasks—especially dense predictive image synthesis—remains largely unexplored. A key challenge is that foundation models primarily encode high-level semantic features, whereas image generation additionally requires accurate local appearance and fine-grained texture. Bridging the “representation-to-generation” gap remains an open problem. 

In addition, conventional pixel-level losses (e.g., L1 or SSIM-based losses) primarily constrain intensity similarity but fail to capture high-level semantic consistency, which may degrade downstream task performance, e.g., segmentation~\cite{perceptual_loss}. To overcome this limitation, perceptual losses introduce feature-level supervision using pretrained networks, enabling the generator to align with deeper semantic representations ~\cite{perceptual_loss}. However, most existing perceptual losses rely on supervised networks such as VGG pretrained on ImageNet~\cite{perceptual_loss_ct}. The VGG feature is not only biased toward classification tasks, but also emphasizes high-level semantic abstraction while lacking pixel-level discriminative ability, making it suboptimal for dense prediction tasks like image generation~\cite{perceptual_loss}. In contrast, the Transformer trained by DINOv3 with Gram anchoring~\cite{dinov3} not only provides less biased features, but also produces more discriminative and semantically rich dense feature maps, which are better aligned with the requirements of pixel-wise regression tasks~\cite{dinov3}.
Therefore, replacing traditional VGG-like backbones with foundation models in the perceptual loss has a potential to enhance image generation quality and anatomical consistency~\cite{dinov3}.

To address these challenges, we propose the DINOv3-Guided Cross-Fusion (DGCF) framework for medical image-to-image translation. DGCF comprises two components: (1) DGCN (DINOv3-Guided Cross-Fusion Network), a hybrid generator that integrates a frozen, self-supervised DINOv3 Transformer with a trainable CNN encoder–decoder via hierarchical cross-fusion to combine global semantic awareness and local structural reconstruction; and (2) MLDP (Multi-Layer DINOv3 Perceptual) loss, a multi-scale perceptual constraint computed in the DINOv3 feature space that enforces semantic and structural consistency between synthesized and real CTs. Experiments on pelvic CBCT$\rightarrow$CT and MRI$\rightarrow$CT translation tasks demonstrate that our method outperforms state-of-the-art methods in both intensity- and semantics-based metrics. To the best of our knowledge, this is the first use of DINOv3 features for medical image generation and opening new avenues for foundation model-guided synthesis in clinical imaging.

\section{Method}

\label{sec:meth}
\begin{figure}
\begin{minipage}{1.0\linewidth}
  \centering
  \centerline{\includegraphics[width=8.8cm]{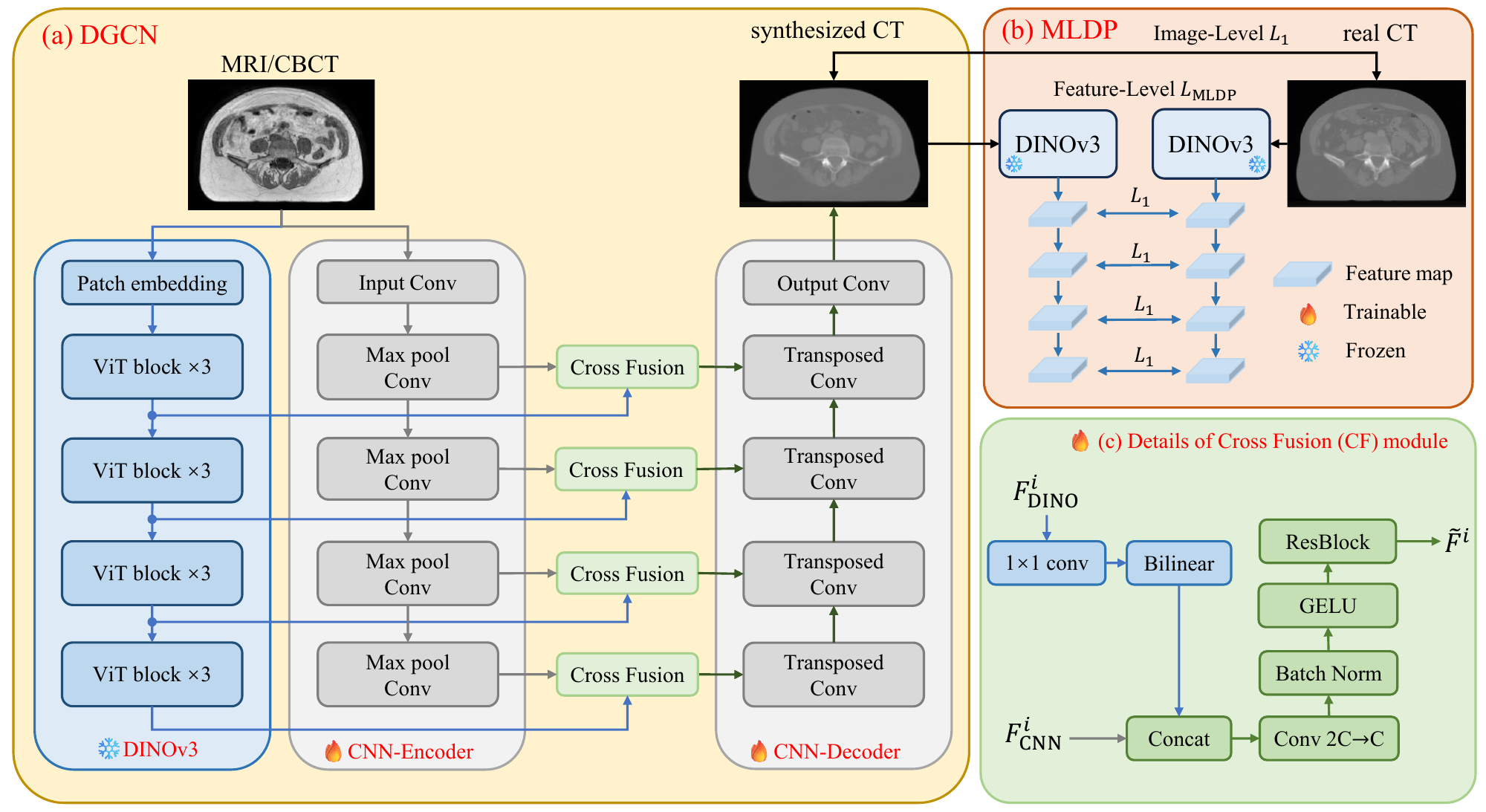}}
\caption{Overview of our DINOv3-Guided Cross Fusion Framework (DGCF) for synthetic CT generation. 
}
\label{fig0}
\end{minipage}
\end{figure}

\begin{figure*}[htbp]
    \centering
    \includegraphics[width=1\linewidth]{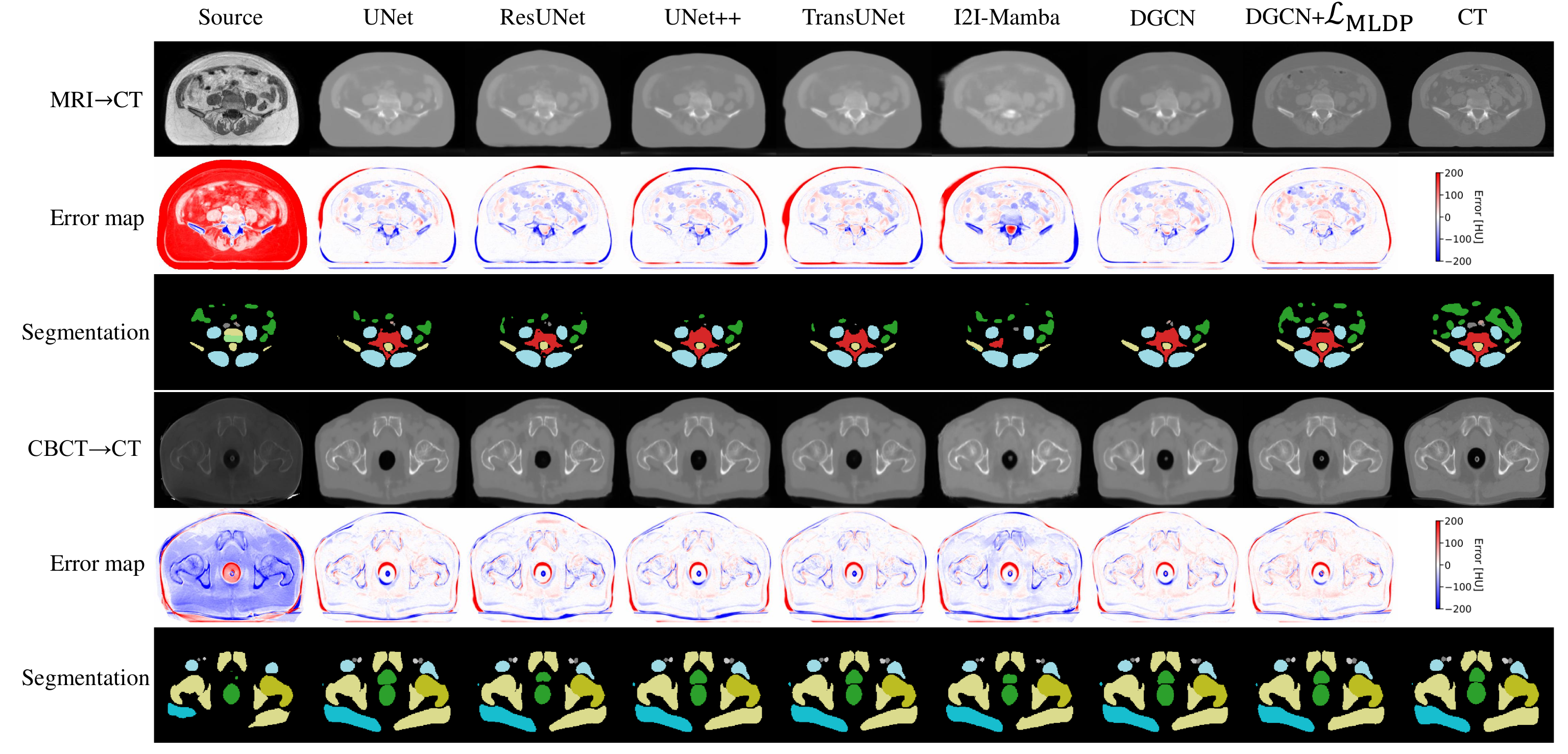}
    \caption{Visual comparison of CT synthesis quality. For each case, the last row shows segmentation obtained by TotalSegmentator from the synthesized images. 
    }
    \label{comparison_figure}
\end{figure*}

\subsection{Overview}

Let $X$ denote an input image in the source modality (e.g., CBCT or MRI), and $Y$ denote the corresponding image in the target modality (e.g., CT). We aim to train a neural network $f_{\theta}$ to obtain synthesized image $Y' = f_{\theta}(X)$ so that $Y'$ is close to $Y$ as much as possible. Our proposed DGCF is illustrated in Fig.~1. It consists of two main components: (1) a DINOv3-Guided Cross Fusion Network (DGCN), which integrates a frozen DINOv3 Transformer encoder with a trainable CNN encoder–decoder through a hierarchical cross-fusion mechanism, and (2) a Multi-Level DINOv3-Perceptual (MLDP) loss that provides semantically meaningful supervision in the feature space of DINOv3 at multiple scales.

\subsection{DINOv3-Guided Cross Fusion Network (DGCN)}

Our proposed DGCN consists of three key components:
(1) a frozen DINOv3 ViT-B/16 encoder~\cite{dinov3} with 12 cascaded Transformer blocks (embedding dimension 768) that extracts hierarchical global semantics while preserving its pretrained feature space to avoid degradation on limited medical datasets;
(2) a trainable CNN encoder following the UNet~\cite{Unet} design to capture spatially precise local structures through cascaded convolutional blocks and max-pooling; and
(3) a symmetric CNN decoder that predicts the synthesized target modality image $Y'$ based on features from the two encoders. 

To effectively integrate the complementary strengths of the two encoders, we design a Cross Fusion (CF) module that aligns and merges multi-level features from DINOv3 and the CNN encoder.
At each corresponding level $i$ ($i = 1,2,3,4$), let $C$ denote the number of channels in the CNN feature map $F_{\text{CNN}}^i$ after the $i$-th convolution block. We use $F_{\text{DINO}}^i$ to denote the feature after the $3i$-th Transformer block in ViT-B, and it is first transformed by a $1{\times}1$ convolution to match the channel dimension $C$, followed by bilinear interpolation to align spatial resolution:
\begin{equation}
F'^i_{\text{DINO}} = \text{Bilinear}\big(\text{Conv}_{1\times1}(F_{\text{DINO}}^i)\big).
\end{equation}

The aligned DINOv3 and CNN features are then concatenated ($\oplus$) and fused through a 3$\times$3 convolution, Batch Normalization (BN) and an activation function based on Gaussian Error Linear Unit (GELU):
\begin{equation}
\tilde{F}^i = \text{GELU}\big(\text{BN}(\text{Conv}_{3\times3}(F_{\text{CNN}}^i \oplus F'^i_{\text{DINO}}))\big).
\end{equation}

Finally, a residual refinement is applied to enhance feature representation:
\begin{equation}
\tilde{F}^i = \tilde{F}^i + \text{Conv}_{3\times3}(\tilde{F}^i).
\end{equation}

Then $\tilde{F}^i$ is sent into the $i$-th resolution level of the CNN decoder that uses transposed convolution to recover local details for CT synthesis. 

\subsection{Multi-Level DINOv3 Perceptual (MLDP) loss}

Traditional VGG-based perceptual losses suffer from limited capability of pixel-level discrimination and task-specific biases. In contrast, DINOv3 has a more local  feature discrimination ability in addition to global awareness with more generalizable features. This enables stronger supervision for CT texture recovery and anatomical consistency. Therefore, we design a perceptual loss that measures feature similarity in the DINOv3 embedding space. Let $Y'$ denote the synthesized CT image and $Y$ the ground-truth CT. Both $Y'$ and $Y$ are fed into the frozen DINOv3 encoder, and the L1 loss is computed between their corresponding feature maps:

\begin{equation}
\mathcal{L}_{\text{MLDP}} = \sum_i \| \phi_i(Y') - \phi_i(Y) \|_1 ,
\end{equation}
where $\phi_i(\cdot)$ represents the feature extracted from the $i$-th layer of DINOv3. In this work, we use features from layers $i = {3, 6, 9, 12}$ to capture multi-level semantic information. The final objective combines pixel-wise reconstruction loss and perceptual loss:
\begin{equation}
\mathcal{L} = \mathcal{L}_{L1} + \lambda \mathcal{L}_{\text{MLDP}},
\end{equation}
where $\mathcal{L}_{L1}=\|Y'- Y\|_1$ measures image-level intensity differences, and $\lambda$ balances the contribution of the perceptual term. Following~\cite{percep_weight}, $\lambda$ is set to 1.0 in our experiments.

\section{EXPERIMENTS AND RESULTS}

\subsection{Data and Implementation}

Our experiments were carried out on the SynthRAD2023 pelvic dataset~\cite{synthrad2023}, covering both the MRI$\rightarrow$CT (Task 1) and CBCT$\rightarrow$CT (Task 2) translation settings, with 180 paired 3D volumes available for each task. Dataset were split into training/validation/test sets in a 7:1:2 ratio. We performed slice-wise translation with slices resized to 256$\times$256. CBCT/CT intensities were clipped to $[-1024, 2000]$ HU and normalized to $[-1, 1]$, while MRI was normalized using its 99th percentile. The model was implemented in PyTorch and trained for 100 epochs on an RTX 2080 Ti using Adam (learning rate=$2\times10^{-4}$, batch size 4). The DINOv3 (ViT-B/16) encoder was frozen, and only the CNN encoder–decoder and cross-fusion module were trainable. The channel number $C$ was set to 64/128/256/512 for the CNN encoder/decoder at different scales. For evaluation, the generated slices were mapped back to HU space and reassembled into 3D volumes at the original resolution.

To comprehensively evaluate the 3D image quality and clinical relevance of the generated sCT, we employ three metrics: Structural Similarity (SSIM)~\cite{ms-ssim&psnr} calculated at multiple scales to assess both fine details and global consistency, Peak Signal-to-Noise Ratio (PSNR)~\cite{ms-ssim&psnr}, and SegScore, which assesses anatomical consistency by segmenting both the synthesized and real CTs using TotalSegmentator~\cite{totalsegmentator} and calculating the Dice coefficient between the corresponding organ masks. In this work, as the 3D volumes were scanned in the pelvic region, SegScore was computed only on pelvic-relevant organs (e.g., urinary bladder and prostate) referring to the organ definitions in~\cite{totalsegmentator}.

\subsection{Comparison with Existing Methods}

We compared our method with both CNN- and Transformer-based architectures. For CNNs, in addition to UNet~\cite{Unet}, we include UNet++~\cite{Unet++} and ResUNet~\cite{resunet} that have achieved top performance in the SynthRAD2023~\cite{synthrad2023} and SynthRAD\allowbreak2025~\cite{synthrad2025} challenges. For Transformer-based approaches, we adopt TransUNet~\cite{transunet}, which combines convolutional encoders with ViT-based global modeling. We also compare with the recent Mamba-based image-to-image framework I2I-Mamba~\cite{I2I-mamba}, which leverages state-space modeling to enhance long-range dependency learning.


\begin{table}[t]
\caption{
Quantitative results of MRI→CT and CBCT→CT translation using different network architectures and loss configurations. $L_{\text{VGG}}$ and $L_{\text{MLDP}}$ represent the perceptual losses based VGG and DINOv3, respectively. The best results for each modality are highlighted in bold. * indicates a statistically significant improvement (p-value $<$ 0.05) over the best existing method according to a paired Student’s \textit{t}-test.
}
\label{comparison_table}
\resizebox{\columnwidth}{!}{
\begin{tabular}{c|c|c|ccc}
\hline
\multirow{2}{*}{\makecell[c]{Source \\ Modality}} & \multirow{2}{*}{Network} & \multirow{2}{*}{Loss} & \multicolumn{3}{c}{Metrics} \\
\cline{4-6}
 & & & SSIM (\%) & PSNR (dB) & SegScore (\%) \\
\hline

\multirow{7}{*}{MRI} 
 & UNet~\cite{Unet} & $L_1$ & 84.67 & 26.43 & 82.62 \\
 & ResUNet~\cite{resunet} & $L_1$ & 84.09 & 26.46 & 82.17 \\
 & UNet++~\cite{Unet++} & $L_1$ & 84.97 & 26.44 & 82.79 \\
 & TransUNet~\cite{transunet} & $L_1$ & 85.29 & 26.56 & 82.75 \\
 & I2I-Mamba~\cite{I2I-mamba} & $L_1$ & 83.30 & 26.11 & 81.95 \\
\cline{2-6}
 & DGCN & $L_1$ & \textbf{85.94*} & 26.71 & 83.08 \\
 & DGCN & $+L_{\text{VGG}}$ & 85.60 & 26.66 & 85.44 \\
 & DGCN & $+L_{\text{MLDP}}$ & 85.83 & \textbf{26.73*} & \textbf{86.19*} \\
\hline

\multirow{7}{*}{CBCT} 
 & UNet~\cite{Unet} & $L_1$ & 88.32 & 27.94 & 85.18 \\
 & ResUNet~\cite{resunet} & $L_1$ & 87.94 & 27.95 & 84.45 \\
 & UNet++~\cite{Unet++} & $L_1$ & 88.50 & 28.06 & 85.55 \\
 & TransUNet~\cite{transunet} & $L_1$ & 88.74 & 28.10 & 85.59 \\
 & I2I-Mamba~\cite{I2I-mamba} & $L_1$ & 88.55 & 28.00 & 85.50 \\
\cline{2-6}
 & DGCN & $L_1$ & \textbf{89.06*} & 28.11 & 85.73 \\
 & DGCN & $+L_{\text{VGG}}$ & 88.92 & 28.15 & 87.04 \\
 & DGCN & $+L_{\text{MLDP}}$ & 88.90 & \textbf{28.16*} & \textbf{87.58*} \\
\hline
\end{tabular}
}
\end{table}

As shown in Table~\ref{comparison_table}, when trained with L1 loss, conventional CNNs such as UNet, ResUNet, and UNet++ yield limited results, with SSIM\textless85\%, PSNR\allowbreak\textless\allowbreak26.5\allowbreak dB and SegScore\textless82.8\% for MRI$\rightarrow$CT translation. Their restricted receptive fields and lack of global modeling hinder the preservation of anatomical structures, resulting in relatively low SegScores ($\approx$82.5\%). TransUNet improves the SSIM to 85.29\% and PSNR to 26.56dB by integrating Transformer-based global modeling, respectively. 
In contrast, under the same L1 loss, DGCN attains the highest SSIM (85.94\%), PSNR (26.71dB) and SegScore (83.08\%) for MRI$\rightarrow$CT, and the corresponding scores are 89.06\%, 28.11dB and 87.58\% for CBCT$\rightarrow$CT, outperforming all the compared methods significantly. 

We then compared different loss functions. As shown in Table 1, using perceptual losses in addition to L1 loss does not affect the SSIM and PSNR scores largely, but significantly improves the SegScores. For MRI$\rightarrow$CT, compared with using L1 only, VGG perceptual loss raised the SegScore from 83.08\% to 85.44\%, and the proposed $L_{\text{MLDP}}$ further improved it to 86.19\%. Likewise, for CBCT$\rightarrow$CT, the scores climbed from 85.73\% to 87.04\% with VGG and reached 87.58\% with $L_{\text{MLDP}}$, showing the advantage of DINOv3-based perceptual loss over the traditional VGG-based one. Representative visual comparisons are provided in Fig.~\ref{comparison_figure}.

\subsection{Ablation Study on Network Architecture}

To validate the effectiveness of our DGCN, we conduct an ablation study comparing four encoder variants: ViT-B/16 only, CNN only, our DGCN with the cross fusion module, and DGCN (concat) that simplely concatenates the DINOv3 and CNN features for the input of the decoder blocks.
The quantitative results are summarized in Table~\ref{ablation}.


\begin{table}[t]
\caption{
Ablation study on the effect of different encoder configurations in the MRI$\rightarrow$CT and CBCT$\rightarrow$CT tasks.
}
\label{ablation}
\resizebox{\columnwidth}{!}{
\begin{tabular}{c|c|ccc}
\hline
\multirow{2}{*}{\makecell[c]{Source \\ Modality}} & \multirow{2}{*}{Method} & \multicolumn{3}{c}{Metrics} \\
\cline{3-5}
 & & SSIM (\%) & PSNR (dB) & SegScore (\%)\\
\hline
\multirow{4}{*}{MRI} 
& ViT-B/16 only & 83.72 & 26.30 & 77.55 \\
& CNN only & 85.11 & 26.60 & 82.39 \\
& DGCN (concat) & 85.62 & 26.62 & 82.62 \\
& DGCN (cross fusion) & \textbf{85.94} & \textbf{26.71} & \textbf{83.08} \\
\hline
\multirow{4}{*}{CBCT} 
& ViT-B/16 only & 87.14 & 27.64 & 80.89 \\
& CNN only & 88.34 & 28.08 & 85.17 \\
& DGCN (concat) & 88.91 & 28.08 & 85.51 \\
& DGCN (cross fusion) & \textbf{89.06} & \textbf{28.11} & \textbf{85.73} \\
\hline
\end{tabular}
}
\end{table}

The ViT-B/16-only model performs the worst (e.g., 83.72\% SSIM) because its $16{\times}$ downsampling loses fine anatomical details. The CNN-only model improves performance (85.11\% SSIM, 82.39\% SegScore) by preserving local textures but still lacks global structure. The DGCN (concat) variant brings only slight gains (85.62\% SSIM, 82.62\% SegScore). In contrast, our DGCN achieves the best results (85.94\% SSIM, 83.08\% SegScore), confirming the benefit of the learnable cross-fusion module.

\section{Conclusion}

We propose a DINOv3–CNN hybrid framework that fuses global semantic features from a frozen DINOv3 Transformer with local representations from a CNN encoder–decoder, guided by DINOv3-based perceptual supervision. This design balances global anatomical consistency and local structural fidelity, producing more realistic CT synthesis from CBCT or MRI. Experiments on pelvic CBCT$\rightarrow$CT and MRI$\rightarrow$CT datasets show consistent gains over existing CNN and Transformer-based methods. In the future, we will extend this framework to other modality pairs such as PET$\rightarrow$CT and CT$\rightarrow$MRI for broader cross-domain applications.

\section{compliance with ethical standards}
This research study was conducted retrospectively using human subject data made available in open access. Ethical approval was not required as confirmed by the license attached with the open-access data.

\section{acknowledgments}
This work was supported by the Natural Science Foundation of Sichuan Province under grant 2025ZNSFSC0455.

\bibliographystyle{IEEEbib}
\bibliography{strings,refs}

\end{document}